\documentclass[10pt,twocolumn,letterpaper]{article}

\usepackage{cvpr} 
\usepackage{times}
\usepackage{epsfig}
\usepackage{graphicx}
\usepackage{amsmath}
\usepackage{amssymb}
\usepackage{booktabs}
\usepackage{multirow}
\usepackage{url} 
\usepackage{pifont}
\newcommand{\cmark}{\ding{51}}

\newcommand{\rev}[1]{\textcolor{black}{#1}}

\begin{document}

\title{IJCB-AFMFR 2026: Competition on Adapting Foundation Models for Face Recognition Using Synthetic Training Data}

\author{
Tahar Chettaoui$^{1,*}$, Guray Ozgur$^{1,2,*}$,
Eduarda Caldeira$^{1,2,*}$, Arturas Nakvosas$^{3,+}$,\\
Hatef Otroshi Shahreza$^{4,+}$
S{\'e}bastien Marcel$^{4,+}$, Rishabh Shukla$^{5,+}$,
Aditya Takkar$^{5,+}$,\\ Rushil Khullar$^{5,+}$,
Lalak Yadav$^{5,+}$
Gourav Gupta$^{6,+}$, Anant Gupta$^{6,+}$,\\
Shiqi Yu$^{7,*}$, Vitomir Struc$^{8,*}$, 
Naser Damer$^{1,2,*}$, Fadi Boutros$^{1,*}$\\
$^{1}$Fraunhofer Institute for Computer Graphics Research IGD, Germany\\
$^{2}$Technical University of Darmstadt, Germany, 
$^{3}$Vilnius University, Lithuania\\
$^{4}$Idiap Research Institute, Switzerland, 
$^{5}$Indian Institute of Technology Jammu, India\\
$^{6}$ArogyaPandit Private Limited, India, 
$^{7}$Southern University of Science and Technology, China \\
$^{8}$University of Ljubljana, Slovenia\\[0.5em]
$^{*}$ Competition Organizer \qquad $^{+}$ Competition Participant\\
{\tt tahar.chettaoui@igd.fraunhofer.de}
}

\maketitle
\thispagestyle{empty}

\begin{abstract}
This paper presents a summary of the Competition on Adapting Foundation Models for Face Recognition Using Synthetic Training Data (AFMFR), held at the 2026 International Joint Conference on Biometrics (IJCB 2026). The competition received a total of eight valid submissions from four distinct teams across two complementary tracks: a Full Data Track, in which participants adapt the CLIP ViT-L/14 foundation model using large-scale synthetic identity data, and a Limited Data Track, designed to reflect more resource-constrained adaptation regimes. All training data was generated  exclusively using IDPERTURB.
Submitted solutions are ranked based on verification and identification performance across a diverse suite of benchmarks, including LFW, CFP-FP, AgeDB-30, CALFW, CPLFW, IJB-B, IJB-C, and TinyFace, using the Borda count method. Fairness evaluation is additionally conducted on the RFW dataset across four demographic groups. The results demonstrate that adaptation of the CLIP foundation model with synthetic training data substantially improves over the off-the-shelf model and, in several cases, surpasses the baseline. Notably, full fine-tuning with Sub-Center ArcFace (DMSTI-Neurotechnology) leads the Full Data Track, while rank-stabilized LoRA adaptation (Idiap-BSP) proves most effective under limited-data conditions.

\end{abstract}

\vspace{-5mm}
\section{Introduction}
\rev{Face recognition (FR) has achieved remarkable progress over the past decade, driven by advances in deep neural network architectures, innovations in training objectives, particularly large-margin softmax losses, and the availability of large-scale annotated face datasets~\cite{Deng_2022,DBLP:conf/cvpr/Kim0L22,wang2018cosfacelargemargincosine}. These developments have enabled FR systems to achieve high performance on several unconstrained and challenging benchmarks and have facilitated their deployment across numerous real-world applications. Despite these advances, the continued development of FR systems remains fundamentally dependent on access to large, diverse, and accurately annotated face datasets. However, acquiring such datasets has become increasingly challenging due to legal, ethical, and privacy concerns surrounding the collection and use of biometric data~\cite{gdpr,lirias3838501,SyntheticFRSurvay}. In recent years, several widely used public datasets, including MS-Celeb-1M~\cite{guo2016ms} and VGGFace2~\cite{DBLP:conf/fgr/CaoSXPZ18}, have been withdrawn or subjected to access restrictions, significantly limiting the availability of large-scale training data. These developments have accelerated research into synthetic face generation as a scalable and privacy-preserving alternative for training FR systems~\cite{DBLP:journals/inffus/MelziTVKRLDMFOZZYZWLTKZDBVGFFMUG24,DBLP:conf/fgr/Otroshi-Shahreza24,DEANDRESTAME2025103099}.}

\rev{Recent advances in generative models, particularly diffusion-based methods, have substantially improved the realism, identity consistency, and diversity of synthetic face images. As a result, synthetic data has evolved from a simple data augmentation technique into a viable alternative to real training data for FR \cite{borsukiewicz2026beyond}. To systematically evaluate this emerging paradigm, the biometric community has organized several benchmarking initiatives dedicated to synthetic-data-driven FR. The FRCSyn challenge at WACV 2024 benchmark~\cite{DBLP:conf/wacv/MelziTVKRLDMFOZZYZWLTKZDBVGFFMUGEORMSK24} established one of the first comprehensive evaluation protocols for assessing synthetic face generation methods and their effectiveness for training FR models. Later, the Synthetic Data for Face Recognition (SDFR) competition~\cite{DBLP:conf/fgr/Otroshi-Shahreza24} compared state-of-the-art synthetic face generation pipelines under a unified evaluation framework. In these competitions, participants trained conventional FR models based on the ResNet-50 backbone from scratch using synthetic training data, enabling a fair comparison of synthetic data generation approaches. More recently, the second edition of the FRCSyn challenge at CVPR 2024 further demonstrated significant progress in synthetic face generation while revealing that a noticeable performance gap between models trained on synthetic and real data still remains~\cite{DEANDRESTAME2025103099}. Collectively, these initiatives established synthetic data as a promising direction for privacy-preserving biometrics while identifying data diversity, identity preservation, and intra-class variation as key factors governing downstream recognition performance. However, they did not investigate the emerging paradigm of adapting large pre-trained vision foundation models, which introduces fundamentally different challenges compared to training conventional FR networks (specifically the ResNet50 architecture) from scratch.}

\rev{Concurrently, computer vision has undergone a fundamental transformation with the emergence of large-scale vision foundation models, such as CLIP~\cite{DBLP:conf/icml/RadfordKHRGASAM21} and DINOv2~\cite{oquab2024dinov2learningrobustvisual}. Unlike conventional FR models that are trained from scratch on face datasets, these models learn rich and transferable visual representations from billions of image-text pairs or self-supervised objectives using internet-scale data. Consequently, they have become the backbone of numerous downstream vision applications through efficient fine-tuning or parameter-efficient adaptation. This paradigm has recently been extended to FR. In particular, FRoundation~\cite{DBLP:journals/ivc/ChettaouiDB25} demonstrated that adapting CLIP \cite{DBLP:conf/icml/RadfordKHRGASAM21} and DINOv2 \cite{oquab2024dinov2learningrobustvisual} using parameter-efficient fine-tuning on synthetic face datasets yields competitive recognition performance, outperforming vision transformers trained from scratch while requiring significantly fewer trainable parameters. These findings suggest that synthetic data can serve not only as a replacement for real training datasets but also as an effective means of adapting powerful pre-trained foundation models to biometric recognition tasks.
Despite these promising developments, systematic and reproducible benchmarking of foundation model adaptation strategies for FR remains largely unexplored. Previous synthetic-data competitions primarily investigated the quality of synthetic images for training conventional FR models from scratch, whereas adapting foundation models introduces fundamentally different research questions concerning parameter-efficient fine-tuning, full-model adaptation, optimization strategies, and the interaction between synthetic identity diversity and pre-trained representations.}

\rev{To address this gap, we introduce the IJCB-AFMFR 2026 Competition on Adapting Foundation Models for Face Recognition Using Synthetic Training Data. Unlike previous synthetic-data competitions \cite{DBLP:conf/fgr/Otroshi-Shahreza24, DBLP:conf/wacv/MelziTVKRLDMFOZZYZWLTKZDBVGFFMUGEORMSK24, DEANDRESTAME2025103099} that focused on training conventional FR models, the proposed competition isolates the adaptation problem by fixing both the foundation backbone and the synthetic data generation pipeline, enabling a fair comparison of different adaptation strategies. Participants are required to adapt the OpenAI CLIP ViT-L/14 foundation model using synthetic face datasets generated exclusively with IDPERTURB~\cite{boutros2026idperturb}, a geometry-driven framework that improves intra-class diversity through identity embedding perturbation. The competition comprises two complementary tracks. The Full Data Track provides approximately three million synthetic images covering 35,000 identities to evaluate large-scale adaptation, whereas the Limited Data Track restricts training to 250,000 images from 5,000 identities, reflecting realistic resource-constrained development scenarios.}

\rev{Submitted methods are evaluated on a comprehensive set of nine FR benchmarks. 
Specifically, evaluation includes the common verification datasets LFW~\cite{huang:inria-00321923}, CFP-FP~\cite{c3517bca662f4193a58fd8f9e3145c8f}, AgeDB-30~\cite{moschoglou2017agedb}, CA-LFW~\cite{DBLP:journals/corr/abs-1708-08197}, and CP-LFW~\cite{CPLFWTech}, the challenging mixed-media benchmarks IJB-B~\cite{inproceedingsijbb} and IJB-C~\cite{DBLP:conf/icb/MazeADKMO0NACG18}, and the low-resolution identification benchmark TinyFace~\cite{DBLP:conf/accv/ChengZG18}. In addition, demographic fairness is assessed using the RFW benchmark~\cite{DBLP:conf/iccv/WangDHTH19}. 
}

\rev{The competition attracted 28 registered teams from academia and industry across multiple countries, resulting in eight valid submissions from four participating teams across the two competition tracks. The submitted methods explore a diverse range of adaptation strategies, including parameter-efficient LoRA fine-tuning~\cite{hu2021loralowrankadaptationlarge}, rank-stabilized LoRA (rsLoRA)~\cite{kalajdzievski2023rank}, and full backbone fine-tuning. Through a comprehensive analysis of recognition performance, demographic fairness, computational considerations, and adaptation strategies, this paper provides the first benchmark of synthetic-data-driven adaptation of foundation models for FR and establishes a reproducible reference for future research in this emerging area.}

\section{Backbone, Training Datasets, Evaluation Setup, and Participants} \label{sec:backbone}

\subsection{Backbone} 
\textbf{CLIP Foundation Model:} 
In this competition, participants are restricted to using the Contrastive Language–Image Pretraining (CLIP) foundation model \cite{DBLP:conf/icml/RadfordKHRGASAM21}, developed by OpenAI. CLIP is a multimodal model trained to learn joint representations of images and text by associating images with their corresponding textual descriptions through contrastive learning, enabling it to capture rich visual semantics grounded in natural language. The model consists of two main components: an image encoder, which maps images into a shared embedding space, and a text encoder, which projects textual descriptions into the same space. This alignment allows the model to measure cross-modal similarity and has demonstrated strong generalization across a wide range of visual recognition tasks, making it a compelling backbone for adaptation. Depending on their approach, participants may leverage both the image and text encoders jointly, or restrict their adaptation to the image encoder alone.

\textbf{Competition Backbone:} 
Among the available CLIP variants, participants are required to use the pretrained ViT-L/14 model, which corresponds to the largest ViT architecture explored in the original CLIP paper \cite{DBLP:conf/icml/RadfordKHRGASAM21}. When using the officially released pre-trained weights provided by the original authors and employing CLIP models purely as frozen feature extractors without additional fine-tuning, ViT-L/14 consistently achieved superior performance across multiple FR benchmarks compared to other CLIP variants \cite{DBLP:journals/ivc/ChettaouiDB25}. The large-scale ViT-L/14 model contains approximately 0.3 billion parameters and also provides a variant, namely ViT-L/14@336, which was additionally pre-trained for one epoch at a higher input resolution of 336 pixels to further improve performance \cite{DBLP:conf/nips/TouvronVDJ19}. However, participants in the competition were restricted to using only the original ViT-L/14 variant operating at the standard 224×224 input resolution employed during the initial training phase \cite{DBLP:conf/icml/RadfordKHRGASAM21}. \rev{The competition intentionally fixed the foundation model (CLIP) and backbone (ViT-L/14), as well as the synthetic dataset used for fine-tuning, to isolate the source of performance differences and ensure a fair, controlled comparison across adaptation strategies, rather than allowing variation in model capacity, pre-training data, or backbone choice to influence the results.
}

\textbf{Baseline:} 
As a baseline, we adopted FRoundation \cite{DBLP:journals/ivc/ChettaouiDB25}, which investigates the adaptation of CLIP \cite{DBLP:conf/icml/RadfordKHRGASAM21} and DINOv2 \cite{oquab2024dinov2learningrobustvisual} foundation models to FR via LoRA \cite{hu2021loralowrankadaptationlarge}, a parameter-efficient fine-tuning mechanism that injects trainable components into the frozen model weights. Their study is directly relevant to our setting: they evaluate the ViT-L/14 CLIP architecture, demonstrate that fine-tuning on synthetic face data improves over both off-the-shelf foundation models and ViTs trained from scratch, and show that this advantage is most pronounced under limited data conditions. FRoundation therefore serves as a strong and principled baseline for our work, as it shares the same backbone, operates in the same low-data synthetic regime, and provides a well-validated fine-tuning protocol for FR. For the competition baseline, we follow the training protocol, hyperparameters, and fine-tuning methodology proposed in FRoundation. The CLIP ViT-L/14 backbone is fine-tuned on the provided synthetic training dataset using the same optimization settings and FR objective described in \cite{DBLP:journals/ivc/ChettaouiDB25}. It follows the same adaptation data as the competition setup, along with all competition constraints and limitations. This baseline therefore serves as a reference implementation for assessing the effectiveness of alternative adaptation strategies and training approaches developed by participants.

\subsection{Synthetic Training Dataset}
The competition is organized into two complementary tracks designed to evaluate adaptation strategies under different data availability regimes. Track 1 focuses on large-scale adaptation, reflecting scenarios where abundant synthetic data can be leveraged to effectively adapt foundation models. Participants are provided with two datasets: a primary dataset containing 2.5 million images spanning 25,000 identities (100 images per identity), and a secondary dataset consisting of 500,000 images across 10,000 identities (50 images per identity). The secondary dataset provides different intra-class variation to enable more possibilities in the adaptation process. In contrast, Track 2 investigates adaptation under limited-data conditions, representing more resource-constrained settings where only a modest amount of synthetic data is available. This track provides a single dataset comprising 250,000 images distributed over 5,000 identities (50 images per identity). Across both tracks, all datasets consist exclusively of synthetic face imagery with identity annotations. Participants must perform model adaptation strictly using the datasets provided for their respective track, and the use of any external data sources is prohibited.

All training datasets provided across both tracks were generated using IDPERTURB \cite{boutros2026idperturb}, a geometry-driven synthetic face generation framework. The choice of this framework is motivated by a key limitation shared by many existing identity-conditioned diffusion models: while capable of producing photorealistic and identity-consistent face images, they often suffer from limited intra-class variation, a critical property for training robust and generalizable FR models. IDPERTURB directly addresses this shortcoming by perturbing identity embeddings within a constrained angular region on the unit hypersphere, producing a diverse set of embeddings without requiring any modification to the underlying generative model. Each perturbed embedding is then used as a conditioning vector for a pre-trained diffusion model, enabling the synthesis of visually varied yet identity-coherent face images. This makes IDPERTURB a principled and scalable solution for large-scale synthetic dataset construction, as it enhances intra-class diversity purely in the embedding space, without relying on auxiliary labels or network-level modifications. Empirically, FR models trained on datasets generated with IDPERTURB \cite{boutros2026idperturb} have demonstrated improved performance across multiple benchmarks compared to existing synthetic data generation approaches, further motivating their adoption in this competition. \ rev {In addition to the SOTA performance achieved by the FR model trained on IDPERTURB \cite{boutros2026idperturb}, as stated in \cite{boutros2026idperturb}, we additionally validated  our choice of IDPERTURB \cite{boutros2026idperturb} by comparing the performances of foundation model, i.e., our CLIP baseline model, finetuned with several SOTA synthetic datasets, including IDiff-Face \cite{DBLP:conf/iccv/BoutrosGKD23}, DCFace \cite{DBLP:conf/cvpr/Kim00023}, Arc2Face \cite{DBLP:conf/eccv/PapantoniouLMDKZ24} and SFace2 \cite{DBLP:journals/tbbis/BoutrosHLSD24} (Table \ref{tab:fr_synth_eval}).}

\subsection{Evaluation Benchmarks} \label{subsec-evalbench}
\textbf{FR evaluations:} We evaluate the performance of the submitted models on several FR benchmarks. These include LFW \cite{huang:inria-00321923}, CFP-FP \cite{c3517bca662f4193a58fd8f9e3145c8f}, AgeDB30 \cite{moschoglou2017agedb}, CA-LFW \cite{DBLP:journals/corr/abs-1708-08197}, and CP-LFW \cite{CPLFWTech}. We report verification accuracies (\%) following the official evaluation protocols for each of these benchmarks. In addition, we evaluated on large-scale evaluation benchmarks, IJB-B \cite{inproceedingsijbb}, and IJB-C \cite{DBLP:conf/icb/MazeADKMO0NACG18}. For IJB-C and IJB-B, we used the official 1:1 mixed verification protocol and reported the verification performance as true acceptance rates (TAR) at false acceptance rates (FAR) of  $1e-3$, $1e-4$, and $1e-5$. These benchmarks were selected because they are commonly used to evaluate the latest advancements in FR and offer a diverse range of use cases \cite{Deng_2022, wang2018cosfacelargemargincosine, ElasticFace, DBLP:conf/iccv/DanLXD0XS23}. We also evaluate on the more challenging identification-based TinyFace \cite{DBLP:conf/accv/ChengZG18} benchmark, which consists of unconstrained, low-resolution face images. Through this evaluation, we assess the model’s robustness on a low-quality face dataset (average 20×16 pixels), highlighting its ability to generalize beyond controlled scenarios. This comprehensive setup enables us to examine the effectiveness of the submitted models across diverse conditions.

\textbf{Bias evaluation:} We evaluate the models on the RFW \cite{DBLP:conf/iccv/WangDHTH19} dataset to assess model bias and performance across demographic groups, as it is widely used for fairness and bias evaluation \cite{DBLP:journals/ivc/ChettaouiDB25, PAMI_BIAS}. The RFW dataset contains four testing subsets corresponding to African, Asian, Caucasian, and Indian groups. We follow the reporting protocols and evaluation metrics associated with the evaluation datasets and recent works \cite{DBLP:journals/ivc/ChettaouiDB25, PAMI_BIAS, DBLP:conf/iccv/WangDHTH19}. We report the results as verification accuracies in (\%) on each subset and as average accuracies to evaluate general recognition performance on the benchmarks. To evaluate the bias, we report the STD between all subsets and the SER, which is given by $\frac{max_gError_g}{min_gError_g}$, where $g$ represents the demographic group, as reported in \cite{PAMI_BIAS, DBLP:journals/corr/abs-1911-10692}. A higher STD value indicates more bias across demographic groups and vice versa. For SER, models with values closer to 1 are less biased.

\begin{table}[t!]
\caption{ A summary of the valid submitted solutions, participating team members, affiliations, and type of institution. More details on the submitted algorithms are provided in Section \ref{sec-solutions}.}
\begin{center} 
\resizebox{\linewidth}{!}{%
\begin{tabular}{l|c|c|c}
\hline
\textbf{Solution} & \textbf{Team Members} & \textbf{Affiliations} & \textbf{Type} \\
\hline
ArogyaPandit & Gourav Gupta, Anant Gupta & Arogyapandit Private Limited, India  & Industry \\ 
\multirow{2}{*}{DMSTI-Neurotechnology} & \multirow{2}{*}{Arturas Nakvosas}                          & Institute of Data Science and Digital Technologies,                               & \multirow{2}{*}{Academia/Industry} \\
                                       &                                                            & \multicolumn{1}{l|}{Vilnius University, Lithuania and Neurotechnology, Lithuania} &                                    \\
Idiap-BSP & Hatef Otroshi Shahreza, S{\'e}bastien Marcel & Idiap Research Institute, Switzerland & Academia \\
STYK IITJ & Rishabh Shukla, Aditya Takkar, Rushil Khullar, Lalak Yadav & Indian Institute of Technology Jammu, India  & Academia \\
\hline
\end{tabular}
}
\end{center}
\label{tab:summary_participants}
\end{table}

\subsection{Evaluation Criteria:}
The competition is divided into two tracks, with teams allowed to submit up to two solutions per track, of which only the best-performing one is considered. Submissions are evaluated and ranked independently per track based on face verification and identification performance across multiple benchmarks. Specifically, verification accuracy is measured on CPLFW, CFP-FP, CALFW, AgeDB30, and LFW, true accept rate (TAR) at multiple false accept rate (FAR) thresholds is measured on IJB-B and IJB-C, and Rank-1 and Rank-5 identification accuracy is measured on TinyFace. For all metrics, higher values indicate better performance. 

Rankings are determined using the Borda count method. For a given metric, all $n$ submitted solutions are ranked by performance, and points are assigned such that the first-placed solution receives $n-1$ points, the second-placed $n-2$, and so on down to $0$ points for the last-placed solution. The final score of each team is computed in three steps. \rev{Following common practice in FR evaluation \cite{DBLP:journals/ivc/ChettaouiDB25, Deng_2022,DBLP:conf/cvpr/Kim0L22, wang2018cosfacelargemargincosine}  and the protocol of previous competition on FR \cite{DBLP:conf/fgr/Otroshi-Shahreza24, DBLP:conf/wacv/MelziTVKRLDMFOZZYZWLTKZDBVGFFMUGEORMSK24, DEANDRESTAME2025103099}, we report single-run results and use an aggregated Borda count across multiple benchmarks to provide a consistent and fair ranking of all submissions.} First, teams are ranked by their average accuracy across the five small verification benchmarks, and a single Borda count is assigned based on this ranking. Second, Borda counts are averaged across the different TAR thresholds of IJB-B and IJB-C separately, then averaged together. Third, Borda counts for Rank-1 and Rank-5 on TinyFace are averaged. The three resulting scores are finally averaged to produce the overall team score, on which the final ranking is based.

As a concrete example, consider $n=4$ teams. A team finishing second on the small benchmarks receives a Borda count of $3$. On IJB-C, if the same team finishes first, second, and third at TAR thresholds $10^{-3}$, $10^{-4}$, and $10^{-5}$ respectively, its IJB-C score is $(4+3+2)/3 = 3$. On TinyFace, finishing first on Rank-1 and third on Rank-5 gives $(4+2)/2 = 3$. The team's final score is then $(3+3+3)/3 = 3$.

\subsection{Competition Participants}
\begin{figure}[!t]
  \centering
   \includegraphics[width=0.7\linewidth]{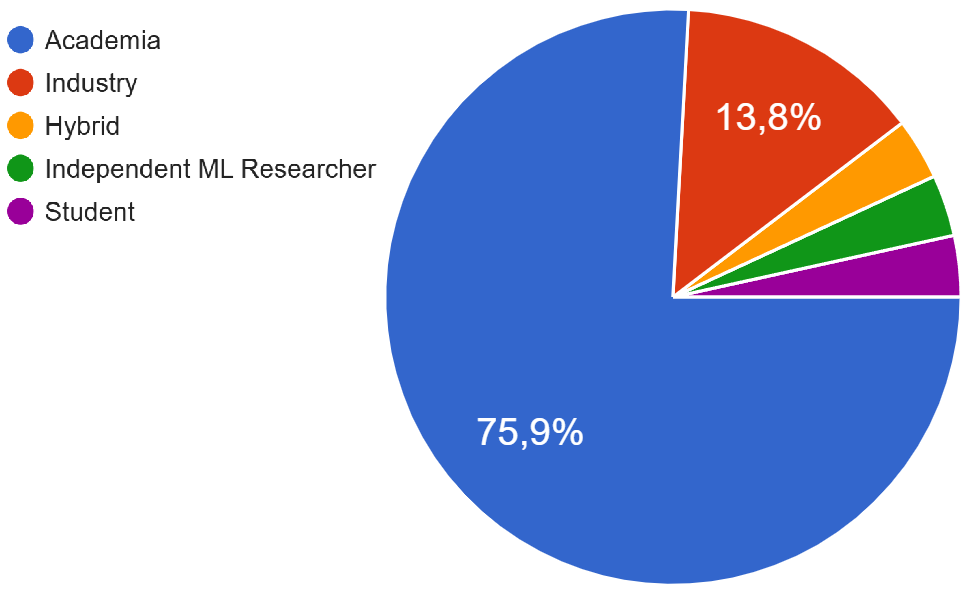}
   \caption{Distribution of registered teams by affiliation type. The competition attracted participants from academia, industry, academia–industry collaborations, independent machine learning researchers, and students, with academic teams accounting for more than 75\% of all registrations.}
   \label{fig:association}
\end{figure}

The competition was designed to attract a broad range of participants from both academia and industry, with diverse geographic and research backgrounds. To maximize outreach, the call for participation was disseminated through the International Joint Conference on Biometrics (IJCB) 2026 website, the competition website, social media platforms, and targeted mailing lists. The competition attracted 28 registered teams from both academia and industry \rev{of which 4 teams submitted 8 valid solutions}.  
Participants represented a diverse range of backgrounds, including academic institutions, industry organizations, academia–industry collaborations, independent machine learning researchers, and students. As shown in Figure \ref{fig:association}, more than 75\% of the registered teams were affiliated with academia. The participants represented a diverse international community, spanning multiple countries across Asia, Africa, Europe, North and South America. Participating organizations included research universities, independent research institutes, and an industrial research startup, reflecting the broad interest in foundation models for FR. Each team was allowed to submit up to two solutions per track, resulting in a total of eight valid submissions across the two competition tracks from four different teams. A summary of the participating teams is provided in Table~\ref{tab:summary_participants}.
\rev{
Although the competition attracted 28 registered teams from academia and industry, only four teams ultimately submitted valid solutions. As in many research competitions, registration reflected initial interest, whereas completing a submission required substantial implementation and computational effort. Participants were required to adapt a large CLIP ViT-L/14 foundation model using large-scale synthetic datasets, comply with the prescribed competition constraints, and submit the final model in the required ONNX format. To facilitate participation, the organizers provided continuous support through the competition website \footnote{
\small{https://sites.google.com/view/ijcb-afmfr-2026/home}}, GitHub repository \footnote{\url{https://github.com/TaharChettaoui/IJCB-AFMFR-2026}}, and email communication. No organizational or technical issues preventing submission were reported to the organizers during the competition.
}

\begin{table*}[ht]
\caption{Summary of the submitted approaches training details across the Full Data and Limited Data tracks. The table reports  the adopted adaptation strategy (e.g., LoRA, rsLoRA, full fine-tuning), the set of adapted transformer components, rank and configuration details where applicable, and the employed loss functions. All methods are built upon the pre-trained CLIP ViT-L/14 backbone.}
\centering
\renewcommand{\arraystretch}{1.3}
\resizebox{\textwidth}{!}{%
\begin{tabular}{lccccc}
\toprule
\textbf{Team} & \textbf{Track} & \textbf{Adaptation Method} & \textbf{Adapted Layers} & \textbf{Rank / Config} & \textbf{Classification Head \& Loss} \\
\midrule

\multirow{2}{*}{Baseline \cite{DBLP:journals/ivc/ChettaouiDB25}}
  & Full Data    & \multirow{2}{*}{rsLoRA \cite{kalajdzievski2023rank}}  & \multirow{2}{*}{$Q, V$} & \multirow{2}{*}{$r{=}16$, $\alpha{=}32$} & \multirow{2}{*}{CosFace ($m{=}0.3$) \cite{wang2018cosfacelargemargincosine}}\\
  & Limited Data & & & & \\
\midrule \midrule

\multirow{2}{*}{ArogyaPandit}
  & Full Data    & \multirow{2}{*}{LoRA \cite{hu2021loralowrankadaptationlarge}}  & \multirow{2}{*}{$Q, V$ } & $r{=}16$, $\alpha{=}32$ & ArcFace ($m{=}0.42$) \cite{DBLP:conf/cvpr/DengGXZ19} \\
  & Limited Data & & & $r{=}4$, $\alpha{=}8$ & ArcFace ($m{=}0.35$) \cite{DBLP:conf/cvpr/DengGXZ19} + embedding-preservation loss \\
\midrule

\multirow{2}{*}{DMSTI-Neurotechnology}
  & Full Data    & Full fine-tuning & \multirow{2}{*}{All layers}  & -- & Sub-Center ArcFace ($K{=}5$, $m{=}0.4$) \cite{DBLP:conf/eccv/DengGLGZ20} \\
  & Limited Data & Full fine-tuning + EMA &  & -- & ArcFace ($m{=}0.4$) \cite{DBLP:conf/cvpr/DengGXZ19} \\
\midrule

\multirow{2}{*}{Idiap-BSP}
  & Full Data    & \multirow{2}{*}{rsLoRA \cite{kalajdzievski2023rank}} & \multirow{2}{*}{$Q, V$} & $r{=}256$, $\alpha{=}16$ & \multirow{2}{*}{CosFace ($m{=}0.3$) \cite{wang2018cosfacelargemargincosine}} \\
  & Limited Data & & & $r{=}1024$, $\alpha{=}16$  &  \\
\midrule

\multirow{2}{*}{STYK IITJ}
  & Full Data    & LoRA \cite{hu2021loralowrankadaptationlarge} & $Q, V$, output proj., LayerNorm  &  \multirow{2}{*}{$r{=}16$, $\alpha{=}32$} & CosFace ($m{=}0.3$) \cite{wang2018cosfacelargemargincosine} \\
  & Limited Data & rsLoRA \cite{kalajdzievski2023rank} & $Q, K, V$, output proj., LayerNorm, FFN proj. & & Sub-Center AdaFace ($K{=}2$, $m{=}0.4$) \cite{DBLP:conf/cvpr/Kim0L22, DBLP:conf/eccv/DengGLGZ20}\\

\bottomrule
\end{tabular}}
\label{tab:approaches}
\end{table*}

\begin{table*}[ht]
\caption{Data pre-processing and augmentation strategies across all submitted teams. Since all methods fine-tune a CLIP ViT-L/14 backbone, all teams uniformly resize input images to $224{\times}224$ pixels and apply standard CLIP normalization. The rightmost column reflect the applied data augmentation during fine-tuning.}
\centering
\renewcommand{\arraystretch}{1.4}
\resizebox{\textwidth}{!}{%
\begin{tabular}{lccc}
\toprule
\textbf{Team} & \textbf{Track}  & \textbf{Resize \& CLIP Normalization} & \textbf{Training Augmentation} \\
\midrule

\multirow{2}{*}{Baseline \cite{DBLP:journals/ivc/ChettaouiDB25}}
  & Full Data  & \multirow{2}{*}{\cmark} & \multirow{2}{*}{Horizontal flipping, and RandAugment \cite{Randaugment_CVPR}} \\
  & Limited Data & &  \\
\midrule \midrule

\multirow{2}{*}{ArogyaPandit}
  & Full Data  & \multirow{2}{*}{\cmark} & \multirow{2}{*}{Random resized crop, horizontal flip, color jitter, random grayscale, Gaussian blur, random erasing} \\
  & Limited Data & &  \\
\midrule

\multirow{2}{*}{DMSTI-Neurotechnology}
  & Full Data   & \multirow{2}{*}{\cmark}  & \multirow{2}{*}{Random resized crop, horizontal flip, color jitter, random grayscale, Gaussian blur, and random erasing} \\
  & Limited Data & & \\

\midrule

\multirow{2}{*}{Idiap-BSP}
  & Full Data  & \multirow{2}{*}{\cmark}  & RandAugment \cite{Randaugment_CVPR}, geometric transforms (shear, translate, rotate), photometric transforms (brightness, color, \\
  & Limited Data & & contrast, sharpness), perceptual transforms (AutoContrast, equalize, grayscale), horizontal flip \\
\midrule

\multirow{2}{*}{STYK IITJ}
  & Full Data  & \multirow{2}{*}{\cmark}  & Horizontal flip, RandAugment \cite{Randaugment_CVPR} \\
  & Limited Data & & Horizontal flip, RandAugment \cite{Randaugment_CVPR}, random affine, random erasing \\

\bottomrule
\end{tabular}}
\label{tab:preprocessing}
\end{table*}

\subsection{Submission and Evaluation Process}
To participate in the competition, teams were first required to submit a registration request including their team name and institutional affiliation, upon which a link to the training data, hosted on a cloud service, was shared via email. Regarding model submission, participants were required to provide their final model in ONNX format, for which official export code was provided through the competition's GitHub repository. The submitted model must strictly adhere to the CLIP ViT-L/14 architecture, and the use of any external models beyond the provided CLIP image and text encoders is explicitly prohibited. This restriction notably excludes the use of pretrained FR models, for instance for knowledge distillation purposes. Additionally, no external setup, installation, or internet access is permitted at runtime. To ensure the integrity and fairness of the evaluation, participating teams were given no access to, or knowledge of, the evaluation benchmark at any point during the competition, including any information regarding the synthetic face generation framework used to produce the training datasets for both tracks.

\section{Submitted Solutions} \label{sec-solutions}
%
A total of 28 teams registered for the competition, reflecting broad interest in adapting foundation models for FR using synthetic training data. To encourage methodological exploration, each team was permitted to submit up to two adapted models per track. In the final evaluation phase, four distinct teams successfully submitted valid solutions for both competition tracks. Solution names, team members, affiliations, and type of the institution, i.e., academic, industry, or mixed, are summarized in Table \ref{tab:summary_participants}. In the following, we provide a brief description of the valid submitted solutions:

\begin{itemize} 

\item \textbf{ArogyaPandit:} The proposed approach adapts the OpenAI CLIP ViT-L/14 model \cite{DBLP:conf/icml/RadfordKHRGASAM21} for FR through LoRA-based \cite{hu2021loralowrankadaptationlarge} fine-tuning of the visual encoder. Specifically, LoRA modules are applied to the \(Q,V\) attention projection layers and optimized using a temporary ArcFace identity classification head trained on the official synthetic identity datasets. For the limited-data track, an additional CLIP embedding-preservation loss is incorporated to improve feature stability under constrained training conditions. During inference preparation, the learned LoRA weights are merged into the original CLIP visual encoder, while the temporary ArcFace classifier is discarded, resulting in an ONNX model that preserves the original CLIP ViT-L/14 visual architecture.

\item \textbf{DMSTI-Neurotechnology:} The proposed approach fine-tunes a CLIP ViT-L/14 backbone~\cite{DBLP:conf/icml/RadfordKHRGASAM21} for face verification. The original contrastive head is replaced with a Sub-Center ArcFace~\cite{DBLP:conf/eccv/DengGLGZ20} classification head in Track~1, configured with \(K=5\), \(s=64\), and \(m=0.4\). For Track~2, a standard ArcFace~\cite{DBLP:conf/cvpr/DengGXZ19} head with \(K=1\) is employed together with an exponential moving average (EMA) mechanism using a decay factor of \(0.99\). The entire backbone is fully unfrozen and optimized using AdamW with cosine learning rate scheduling and warmup, employing a backbone learning rate of \(2 \times 10^{-5}\). Training is performed in bf16 precision on 8 NVIDIA H100 GPUs with extensive data augmentation, including RandAugment, RandomErasing, and GaussianBlur. During inference, the classification head is discarded, and cosine similarity is computed directly from the \(\ell_2\)-normalized embeddings.

\item \textbf{Idiap-BSP:} The proposed method fine-tunes the image encoder of a pre-trained CLIP ViT-L/14 model \cite{DBLP:conf/icml/RadfordKHRGASAM21} for FR using the rsLoRA technique \cite{kalajdzievski2023rank}, a widely adopted strategy for efficient foundation model adaptation~\cite{shahreza2025foundation}. rsLoRA is applied within the self-attention blocks of the vision transformer using a rank of $r{=}256$ for Track 1 and $r{=}1024$ for Track 2, with a scaling factor of $\alpha{=}16$ and a dropout rate of 0.25. The adapted image encoder is coupled with a CosFace classification head~\cite{wang2018cosfacelargemargincosine} configured with s=64 and m=0.3, and the entire framework is trained end-to-end using a cross-entropy objective over margin-modified cosine logits. Optimization is performed using AdamW with cosine annealing learning rate scheduling and gradient clipping at $\ell_2$ norm $5$. Training is conducted for 20 epochs on the full training set of each track, using a batch size of 368 on 8 NVIDIA RTX 3090 GPUs for Track 1, and a batch size of 184 on 4 NVIDIA RTX 3090 GPUs for Track 2.

\item \textbf{STYK IITJ:} The proposed approach adapts a pre-trained CLIP ViT-L/14 \cite{DBLP:conf/icml/RadfordKHRGASAM21} image encoder for masked face verification using LoRA \cite{hu2021loralowrankadaptationlarge} while keeping the majority of the backbone parameters frozen. In Track~1, LoRA modules are injected into the query (\(Q\)), value (\(V\)), and output projection layers using a rank of \(r=16\) with \(\alpha/r\) scaling, and optimization is performed using CosFace-based classification losses. For Track~2, the adaptation strategy is extended by additionally applying adapters to the key (\(K\)) and feed-forward network projection layers, combined with rank-stabilized LoRA (rsLoRA) \cite{kalajdzievski2023rank} using \(\alpha/\sqrt{r}\) scaling. This track further incorporates Sub-Center AdaFace with \(K=2\) sub-centers and layer-wise learning rate decay with \(\gamma=0.85\). Across both tracks, LayerNorm parameters are unfrozen during training, a CLIP text-anchor regularization term is employed to preserve semantic consistency, and an SVD-based intra-modal projection method (IsoCLIP) \cite{Magistri_2026_CVPR} is applied during inference.

\end{itemize}

Table~\ref{tab:approaches} summarizes the architectural and training choices of all submitted methods, while Table~\ref{tab:preprocessing} reports the corresponding preprocessing and data augmentation pipelines. These summaries provide a consolidated view of the design decisions explored by participants for adapting the CLIP ViT-L/14 backbone to FR using the provided synthetic training data.

\begin{table*}[ht]
\caption{The achieved verification performances by the baseline and submitted models. The results are reported in (\%) on the small benchmarks, defined in Section \ref{subsec-evalbench}, and as average accuracies. On IJB-B and IJB-C, the results are reported as TAR at FAR of 1e-3, 1e-4 and 1e-5. DMSTI-Neurotechnology achieves the best overall verification performance in the Full Data Track, while Idiap-BSP demonstrates superior robustness on TinyFace and emerges as the most consistent method under limited-data constraints.}
\resizebox{\linewidth}{!}{%
\centering
\begin{tabular}{cc|cccccc|cccccc|cc}
\toprule
\multirow{3}{*}{} &  & & & &  &  &  &  &  &  & & \\
Track & Approach & LFW & CFP-FP & AgeDB30 & CALFW & CPLFW & Avg. &\multicolumn{3}{c}{IJB-B} & \multicolumn{3}{c}{IJB-C} & \multicolumn{2}{c}{TinyFace} \\
& &  & &  & & & & $10^{-3}$ & $10^{-4}$ & $10^{-5}$ & $10^{-3}$ & $10^{-4}$ & $10^{-5}$ & Rank-1 & Rank-5  \\ \hline
- & CLIP & 95.90 & 90.66 & 79.82 & 83.10 & 82.73 & 86.44 & 62.72 & 40.90 & 20.52 & 64.74 & 44.69 & 25.23 & 34.71 & 43.96\\
\hline \hline
\multirow{5}{*}{Full Data Track} & Baseline \cite{DBLP:journals/ivc/ChettaouiDB25} &  99.17	&95.94&	90.68&	91.82&	90.47&	93.62 & 91.18	&82.57 	&69.25 & 93.26 &	86.39	& 76.71& 62.79 & 68.96\\
& Projection Layer & 98.40	&95.46	& 85.28 & 88.48 & 89.22 & 91.37 & 85.85 & 70.69 & 49.01  & 87.26 & 73.39 & 55.83 & 44.64 &54.94 \\
& ArogyaPandit &  95.50	& 92.20	 & 79.90	&81.72	&82.32	&86.33 & 62.77	& 40.50	& 19.98 &64.84	& 44.38	& 24.88 & 34.44 &	43.56 \\
& DMSTI-Neurotechnology & 99.47	&97.39	&94.43	&94.40	&91.88	&\textbf{95.51} & \textbf{94.19}	&\textbf{89.68}	&\textbf{80.52}  &\textbf{95.62}	&\textbf{92.48}	&\textbf{87.42} &65.18	&70.25 \\
& Idiap-BSP &  99.35	&96.86&	92.97&	93.07	&90.98	&94.64 & 92.96&	86.58&	74.18 &94.59	&89.83	&82.46 &\textbf{65.50}	&\textbf{71.14}\\
& STYK IITJ &  99.43	& 96.31	& 90.78	& 92.20	& 90.83	& 93.91 & 92.18& 	83.84	&70.31 & 94.23&	88.04& 	78.06&  61.88&	67.97 \\
\hline
\multirow{5}{*}{Limited Data Track} 
& Baseline \cite{DBLP:journals/ivc/ChettaouiDB25} & 99.10&	96.57&	91.85&	92.60&	90.65&	94.15 & 91.48 &	83.12	& 67.61 & 93.82	& 87.77	& 76.50 & 61.05	& 67.54 \\
& Projection Layer & 98.88&	95.43	&85.15	&89.55&	89.58	&91.72 & 87.23&	70.88	&43.27& 88.68 & 74.59 & 53.08 & 43.80 & 53.031\\
& ArogyaPandit & 99.32	& 95.10	& 84.88	&90.10	&89.73	&91.83 & 87.26	&74.15	&54.40 &89.84	&78.91	&65.65 & 54.94&	62.55 \\
& DMSTI-Neurotechnology &  99.35	&96.46	&91.02	&92.62	&90.95	&94.08 &91.68	&82.40	&54.65& 93.67	&86.08&	60.49  &58.99&	65.91 \\
& Idiap-BSP &  99.47	&96.51	&92.53	&93.53	&90.53	&\textbf{94.52} & \textbf{92.56}	& \textbf{86.23}	& \textbf{74.72} & \textbf{94.51}	& \textbf{89.79}	& \textbf{81.13 }& \textbf{62.90	}& \textbf{68.78}\\
& STYK IITJ &  95.52	& 92.20	& 79.88	& 81.82	& 82.35	& 86.35 & 63.33	& 41.09	& 20.15 & 65.34	& 45.03	& 25.25 & 34.44	& 43.62 \\
\bottomrule
\end{tabular}}
\label{tab:fr_eval}
\end{table*}


\section{Results} \label{sec-results}
This section presents a comprehensive evaluation of all submitted approaches against the FRoundation baseline \cite{DBLP:journals/ivc/ChettaouiDB25}. First, we assess standard FR performance on a diverse set of benchmarks, as described in Section \ref{subsec-evalbench}. Second, we conduct a demographic fairness evaluation to measure the extent to which each approach mitigates bias across ethnic groups. Results are reported separately for the Full Data and Limited Data tracks, allowing a direct assessment of how data availability influences both recognition performance and fairness. CLIP \cite{DBLP:conf/icml/RadfordKHRGASAM21} is included across all evaluations as a reference point to contextualize the gains brought by fine-tuning.

 \begin{table*}[!t]
 \caption{Evaluation on RFW reported as average recognition performance in (\%), standard deviation (STD) and skewed error ratio (SER) across four different demographic groups, as defined in Section \ref{subsec-evalbench}. The higher STD indicates a more biased model and the higher average (avg) indicates, in general, better recognition performance. For SER, models with values closer to 1 are less biased. DMSTI-Neurotechnology leads the Full Data Track in both accuracy and fairness, while Idiap-BSP is the top performer in the Limited Data Track.}
\begin{center}
\resizebox{0.7\linewidth}{!}{%
\renewcommand{\arraystretch}{1.3} 
\begin{tabular}{cc|ccccccc}
\toprule
\textbf{Track} & \textbf{Approach} & African & Asian & Caucasian & Indian & Avg. & STD & SER \\
\midrule
- & CLIP \cite{DBLP:conf/icml/RadfordKHRGASAM21} & 70.75 & 69.73 & 79.32 & 68.98 & 72.19 & 4.80 & 1.49 \\  
\hline \hline
\multirow{5}{*}{Full Data Track} & Baseline \cite{DBLP:journals/ivc/ChettaouiDB25} & 85.23&	85.42&	91.52&	86.12&	87.07&	2.99&	1.74\\
& ArogyaPandit & 74.03	& 72.15	& 82.60	& 73.15	& 75.48	& 4.81	& \textbf{1.60} \\
& DMSTI-Neurotechnology & 90.82	&89.78	&95.08	&91.13	&\textbf{91.70}	&\textbf{2.33}	&2.08 \\
& Idiap-BSP & 87.38	&87.97	&93.25	&88.97	&89.39	&2.65	&1.87 \\
& STYK IITJ & 84.42 &	84.90	&92.33	&87.03	&87.17	&3.62	&2.03 \\
\hline
\multirow{5}{*}{Limited Data Track} & Baseline \cite{DBLP:journals/ivc/ChettaouiDB25} & 84.48&85.77	&91.72	&86.52	&87.12&	3.18&	1.87\\
& ArogyaPandit & 79.78	&81.25	&87.70	&80.82	&82.39	&3.59	&1.64 \\
& DMSTI-Neurotechnology & 84.87 &	86.20	&91.40	&87.25	&87.43	&\textbf{2.82}	&1.76 \\
& Idiap-BSP & 86.60	& 87.38	& 93.18	& 88.52	& \textbf{88.92} & 2.95	&1.97 \\
& STYK IITJ & 74.27	& 72.15	&82.60	&73.17	&75.55	&4.78	&\textbf{1.60}\\
\bottomrule
\end{tabular}}
\end{center}
\label{tab:bias_eval_table}
\end{table*}

\subsection{FR Evaluation} \label{subsec-fr}
To evaluate FR performance across the submitted approaches, we report verification results on LFW \cite{huang:inria-00321923}, CFP-FP \cite{c3517bca662f4193a58fd8f9e3145c8f}, AgeDB-30 \cite{moschoglou2017agedb}, CALFW \cite{DBLP:journals/corr/abs-1708-08197}, and CPLFW \cite{CPLFWTech}. We also report TAR at varying FAR thresholds on the large-scale benchmarks IJB-B \cite{inproceedingsijbb} and IJB-C \cite{DBLP:conf/icb/MazeADKMO0NACG18}, as well as on the more challenging TinyFace benchmark \cite{DBLP:conf/accv/ChengZG18}. The pre-trained CLIP ViT-L/14 \cite{DBLP:conf/icml/RadfordKHRGASAM21} model is included for reference, highlighting that fine-tuning is necessary to achieve competitive performance. The FRoundation \cite{DBLP:journals/ivc/ChettaouiDB25} baseline establishes the competitive benchmark for the Full and Limited Data tracks, respectively. Based on the reported results in Table \ref{tab:fr_eval}, we make the following observations:

\begin{itemize}
    \item In the Full Data Track, DMSTI-Neurotechnology achieves the strongest overall performance, obtaining the best average accuracy on the small-scale benchmarks (95.51\%) and consistently leading on IJB-B and IJB-C across all FAR thresholds. Notably, it substantially surpasses the FRoundation baseline, particularly under strict operating conditions (e.g., 87.42\% TAR@$10^{-5}$ on IJB-C versus 76.71\% for FRoundation), demonstrating superior robustness in low-FAR regimes. Idiap-BSP follows closely with strong overall performance on the small-scale and large-scale benchmarks and achieves the best results on TinyFace (65.50\% Rank-1 and 71.14\% Rank-5), indicating stronger robustness to low-resolution FR. STYK IITJ remains competitive on the small benchmarks (93.91\%) but exhibits a noticeable performance drop at stricter FAR thresholds, suggesting weaker generalization under challenging verification settings.
    
    \item In the Limited Data Track, Idiap-BSP emerges as the most effective and consistent approach, achieving the best average accuracy on small-scale benchmarks (94.52\%) while outperforming the FRoundation baseline across all reported benchmarks. In particular, it achieves substantial gains on IJB-B and IJB-C at low FAR thresholds, highlighting strong robustness despite the reduced training data regime. DMSTI-Neurotechnology remains competitive on the small benchmarks (94.08\%) but experiences a significant degradation at stricter FAR thresholds, especially on IJB-C (60.49\% TAR@$10^{-5}$ compared to 76.50\% for FRoundation), revealing reduced stability under limited-data constraints. 

    \item \rev{Beyond overall ranking, the results reveal a clear trade-off between adaptation strategies. Full fine-tuning (DMSTI-Neurotechnology) achieves the best performance in the Full Data Track, including at strict FAR thresholds, but its performance collapses under the Limited Data Track, indicating that unconstrained updates to the full backbone require sufficient data to be effective and overfit when data is scarce. LoRA-based approaches, in particular Idiap-BSP's high-rank rsLoRA configuration, are more conservative and remain the most consistent method across both tracks, indicating that constrained adaptation offers greater robustness under limited-data conditions. This trade-off echoes prior findings that full fine-tuning tends to outperform LoRA on harder, more data-intensive tasks, while LoRA acts as a stronger regularizer against overfitting when data is limited~\cite{DBLP:journals/tmlr/BidermanPOPGJKH24}.}
    
    \item \rev{Increasing the LoRA rank from the baseline's rank-16 to Idiap-BSP's rsLoRA (rank-256 in the Full Data Track, rank-1024 in the Limited Data Track) yields consistent gains in both tracks: average accuracy on small benchmarks improves from 93.62\% to 94.64\% in the Full Data Track and from 94.15\% to 94.52\% in the Limited Data Track, with larger gains at strict FAR thresholds (IJB-C TAR@1e-5: 76.71\% to 82.46\% Full Data Track, 76.50\% to 81.13\% Limited Data Track). This indicates that rank capacity is a key driver of adaptation quality.}

    \item \rev{In addition to the submitted participants' solutions, we introduce a minimal adaptation baseline, in which only the final projection layer is fine-tuned, while the rest of the backbone remains frozen. This already improves substantially over zero-shot CLIP, but the gap with the LoRA-based baseline widens at stricter FAR thresholds, showing that adapting only the final projection is not enough to reach the fine-grained discriminability needed at low false accept rates.}
\end{itemize}

Based on the evaluation criteria, the top three teams for each track are as follows. In the Full Data Track, DMSTI-Neurotechnology ranks first, achieving the highest average accuracy on the small benchmarks and dominating across all IJB-B and IJB-C TAR thresholds. Idiap-BSP ranks second, leading on TinyFace Rank-1 and Rank-5 while remaining competitive across all other benchmarks. STYK IITJ ranks third, consistently outperforming ArogyaPandit across verification and identification metrics. In the Limited Data Track, Idiap-BSP ranks first, achieving the best performance across all benchmarks including the small verification sets, all IJB-B and IJB-C TAR thresholds, and both TinyFace metrics. DMSTI-Neurotechnology ranks second, performing comparably to Idiap-BSP on the small benchmarks while showing a notable drop at stricter FAR thresholds on IJB-B and IJB-C. ArogyaPandit ranks third, outperforming STYK IITJ across all metrics by a significant margin.

\subsection{Bias Evaluation} \label{subsec-bias}
To assess demographic fairness across submitted approaches, we evaluate each method on the RFW benchmark~\cite{DBLP:conf/iccv/WangDHTH19}, which partitions the test set into four ethnic groups: African, Asian, Caucasian, and Indian. Results are reported in Table~\ref{tab:bias_eval_table} in terms of average recognition accuracy, STD, and SER.  CLIP (72.19\%, STD 4.80) is included for reference, highlighting the substantial gain brought by fine-tuning. The FRoundation \cite{DBLP:journals/ivc/ChettaouiDB25} baseline sets the competitive benchmark at 87.07\% (STD 2.99) and 87.12\% (STD 3.18) for the Full and Limited Data tracks respectively. From the reported results, we make the following observations:

\begin{itemize}
    \item In the Full Data Track, DMSTI-Neurotechnology surpasses the FRoundation baseline \cite{DBLP:journals/ivc/ChettaouiDB25} and all competing approaches by a clear margin, achieving the best accuracy (91.70\%) and lowest STD (2.33), improving both performance and fairness simultaneously. Idiap-BSP ranks second, achieving 89.39\% (STD 2.65). STYK IITJ matches FRoundation in accuracy ($\sim$87\%) but exhibits higher STD (3.62), indicating greater demographic disparity despite comparable average performance. ArogyaPandit falls below the baseline and other competing approaches in accuracy (75.48\%) despite achieving the best SER (1.60), reflecting a trade-off between error balance and overall performance that limits its practical competitiveness.

    \item In the Limited Data Track, rankings shift notably. Idiap-BSP leads all participants (88.92\%, STD 2.95), exceeding competing solutions and the FRoundation baseline \cite{DBLP:journals/ivc/ChettaouiDB25} in both accuracy and fairness. DMSTI-Neurotechnology ranks second, achieving 87.43\% (STD 2.82), and retains the lowest STD in this track, maintaining competitive fairness despite a drop relative to its Full Data performance.

    \item Across both tracks and all methods, the Caucasian group consistently yields the highest per-group accuracy, while the African and Asian groups systematically score lowest, reflecting demographic biases inherited from CLIP pre-training that fine-tuning on synthetic identity datasets only partially mitigates.
\end{itemize}

Overall, DMSTI-Neurotechnology represents the strongest submission in the Full Data Track, excelling in both recognition accuracy and demographic fairness, while Idiap-BSP emerges as the most competitive participant in the Limited Data Track, demonstrating generally robust and fair adaptation under constrained training conditions.

\begin{table*}[ht]
\caption{\rev{The achieved verification performances by the baseline approach, defined in Section \ref{sec:backbone}, fine-tuned on different SOTA synthetic face datasets. We additionally report a real-data baseline trained on CASIA-WebFace, matched in scale to the synthetic datasets (10K identities, 0.5M images), to contextualize these results with respect to real-data training and prior work. The results are reported in (\%) on the small benchmarks, defined in Section \ref{subsec-evalbench}, and as average accuracies. On IJB-B and IJB-C, the results are reported as TAR at FAR of 1e-3, 1e-4 and 1e-5. IDPerturb consistently achieves the best performance across all benchmarks, demonstrating the effectiveness of the proposed synthetic data generation approach compared with prior state-of-the-art synthetic face datasets.}}
\resizebox{\linewidth}{!}{%
\centering
\begin{tabular}{c|cccccc|cccccc}
\toprule
\multirow{3}{*}{} &  & & & &  &  &  &  &  &  & \\
Training Dataset & LFW & CFP-FP & AgeDB30 & CALFW & CPLFW & Avg. &\multicolumn{3}{c}{IJB-B} & \multicolumn{3}{c}{IJB-C}  \\
 &  & &  & & & & $10^{-3}$ & $10^{-4}$ & $10^{-5}$ & $10^{-3}$ & $10^{-4}$ & $10^{-5}$ \\ \hline
\rev{CASIA-WebFace} \cite{DBLP:journals/corr/YiLLL14a} & 99.55 &95.73 &91.73 &92.58 &91.70 &94.26 &90.33 &78.71 &56.96 &92.59 &83.12 &68.87\\
\rev{WebFace4M} \cite{zhu2021webface260mbenchmarkunveilingpower} & 99.65 &96.50 &93.72 &94.37 &93.73 &95.59& 95.12 &90.72 &82.76 &96.62 &93.40 &88.55\\ \hline

\rev{IDiff-Face }(ICCV 2023) \cite{DBLP:conf/iccv/BoutrosGKD23} & 98.75 &91.40 &86.50 &90.78& 84.87 &90.46& 82.92&64.56 &34.82& 86.07& 71.77 &48.18 \\
\rev{DCFace} (CVPR 2023)\cite{DBLP:conf/cvpr/Kim00023} & 98,62 & 94,07	& 90,45	& 91,33	& 87,10	& 92,31 &  86.36  & 73.27 &	51.65  & 89.38 & 77.63  & 56.13  \\
\rev{Arc2Face} (ECCV 2024) \cite{DBLP:conf/eccv/PapantoniouLMDKZ24} & 98,88	& 95,19	& 85,68	& 88,70	& 88,22	& 91,33 & 84.62	& 69.06	& 50.51 & 88.36	& 74.98 & 59.29\\
\rev{SFace2} (TBIOM 2024) \cite{DBLP:journals/tbbis/BoutrosHLSD24} & 97.88 &89.87 &79.03& 86.50& 83.88 &87.43& 75.31& 54.86& 25.14& 78.64 &60.81 &37.73 \\
\rev{IDperturb} (CVPR 2026)\cite{boutros2026idperturb} & 99.37	&96.76	&92.05	&92.92	&90.917	&\textbf{95.27} & \textbf{92.04} 	& \textbf{84.01} & \textbf{67.25} & \textbf{94.07}	& \textbf{88.24}	&\textbf{76.72}\\
\bottomrule
\end{tabular}}
\label{tab:fr_synth_eval}
\end{table*}

\subsection{\rev{Evaluation of Synthetic Face Datasets}} \label{subsec-bias}
\rev{To justify the choice of the synthetic data generation method adopted for the competition, we fine-tune the baseline model \cite{DBLP:journals/ivc/ChettaouiDB25} on several SOTA synthetic face datasets, including IDiff-Face \cite{DBLP:conf/iccv/BoutrosGKD23}, DCFace \cite{DBLP:conf/cvpr/Kim00023}, Arc2Face \cite{DBLP:conf/eccv/PapantoniouLMDKZ24}, SFace2 \cite{DBLP:journals/tbbis/BoutrosHLSD24}, and the IDPerturb dataset \cite{boutros2026idperturb}, which was used to generate the synthetic training data for both the Full Data and Limited Data Tracks. All considered synthetic datasets contain the same number of samples, consisting of 0.5 million synthetic face images generated from 10K identities. Table~\ref{tab:fr_synth_eval} summarizes the verification performance on the considered FR evaluation benchmarks defined in Section \ref{subsec-evalbench}.The baseline model fine-tuned with IDPerturb consistently outperforms models fine-tuned on the other synthetic datasets across all considered evaluation benchmarks. These results provide empirical justification for adopting IDPerturb as the synthetic data generation method for the competition.}

\rev{To contextualize these results with respect to real-data training and prior work, we additionally report a real-data baseline trained on CASIA-WebFace \cite{DBLP:journals/corr/YiLLL14a}, matched in scale to the synthetic datasets used in this competition (10K identities, 0.5M images). In the previous SDFR 2024 competition \cite{DBLP:conf/fgr/Otroshi-Shahreza24}, models trained on synthetic data consistently underperformed relative to real-data training, leaving a substantial synthetic-to-real gap. In contrast, we observe that the baseline fine-tuned on the IDperturb dataset achieves better results than the model trained on CASIA-WebFace at comparable scale. This suggests that adapting foundation models for synthetic data generation is a promising direction for narrowing the synthetic-to-real gap identified in SDFR 2024.}


\begin{figure}[!t]
  \centering
   \includegraphics[width=1\linewidth]{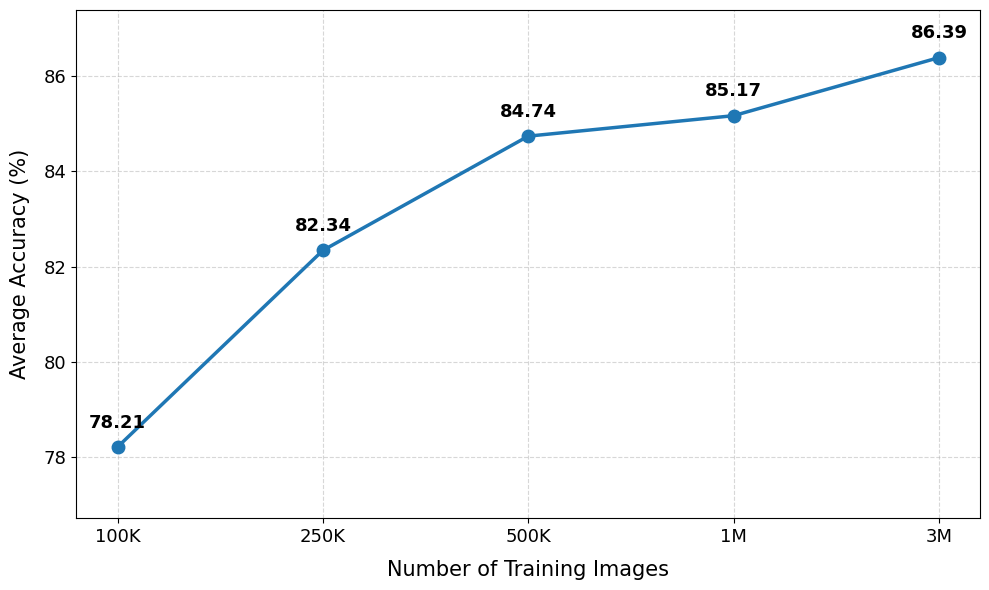}
   \caption{\rev{The effect of data volume on downstream FR performance. The y-axis reports IJB-C TAR at FAR of 1e-4, and the x-axis reports the size of the synthetic dataset subset used to fine-tune the baseline model (100K, 250K, 500K, 1M, and 3M images). Performance improves consistently with dataset size}}
   \label{fig:data_eff}
\end{figure}

\subsection{\rev{Data Efficiency Analysis}} \label{subsec-dataeff}
\rev{To assess how performance changes with the amount of training data, we fine-tuned the baseline model on subsets of 100K, 250K, 500K, 1M, and 3M images and report IJB-C TAR at FAR of 1e-4 in Figure \ref{fig:data_eff}. Performance improves consistently with dataset size, from 78.21\% at 100K to 86.39\% at 3M images, with the largest gains observed between 100K and 500K images (78.21\% to 84.74\%), while further scaling to 1M and 3M images brings smaller additional improvements (85.17\% and 86.39\%, respectively).}

\subsection{\rev{Computational Cost}} \label{subsec-computcost}
\rev{Table \ref{tab:approaches} reports the number of trainable parameters for each submitted approach. The full CLIP ViT-L/14 model, including both image and text encoders, comprises 427,616,513 parameters. Although participants were permitted to leverage the text encoder in their solutions, all submitted approaches relied solely on the image encoder, which accounts for 303,966,208 parameters. Within this image encoder, the number of trainable parameters varies substantially depending on the adaptation strategy, ranging from as few as 0.8M for ArogyaPandit's rank-8 LoRA configuration in the Limited Data Track, to 304M for DMSTI-Neurotechnology's full fine-tuning, which updates the entire image encoder. Idiap-BSP's rsLoRA configurations fall in between, using 25M trainable parameters in the Full Data Track (r=256) and 101M in the Limited Data Track (r=1024). Despite these differences in trainable parameters and training cost, all submitted models share the same CLIP ViT-L/14 image encoder architecture at inference time, resulting in identical inference cost 84.372 GFLOPs across submissions.}

\begin{table}[ht]
\caption{\rev{Number of trainable parameters for  submitted approach. While all methods share the same CLIP ViT-L/14 backbone at inference time, the number of trainable parameters vary depending on the adaptation strategy and the chosen hyperparameters.}}
\centering
\resizebox{0.5\textwidth}{!}{%
\begin{tabular}{lccc}
\toprule
\textbf{Team} & \textbf{Track} & \textbf{Adaptation Method} & \textbf{\#Trainable Parameters(M)} \\
\midrule

\multirow{2}{*}{\rev{Baseline} \cite{DBLP:journals/ivc/ChettaouiDB25}}
  & Full Data    & \multirow{2}{*}{LoRA, r=16}  & \multirow{2}{*}{1.6} \\ 
  & Limited Data & &  \\
\midrule \midrule

\multirow{2}{*}{\rev{ArogyaPandit}}
  & Full Data    & LoRA, r=16  &  1.6 \\ 
  & Limited Data & LoRA, r=8 & 0.8 \\ 
\midrule

\multirow{2}{*}{\rev{DMSTI-Neurotechnology}}
  & Full Data  & \multirow{2}{*}{Full fine-tuning} & \multirow{2}{*}{304} \\ 
  & Limited Data & & \\ 
\midrule

\multirow{2}{*}{\rev{Idiap-BSP}}
  & Full Data    & LoRA, r=256 &  25 \\ 
  & Limited Data & LoRA, r=1024 & 101 \\ 
\midrule

\multirow{2}{*}{\rev{STYK IITJ}}
  & Full Data    & \multirow{2}{*}{LoRA, r=16} & \multirow{2}{*}{1.6}\\ 
  & Limited Data & & \\

\bottomrule
\end{tabular}}
\label{tab:approaches}
\end{table}

\section{Conclusion}
This paper presented the IJCB-AFMFR 2026 Competition on Adapting Foundation Models for FR Using Synthetic Training Data. The competition provided a controlled and reproducible benchmark for evaluating adaptation strategies for the CLIP ViT-L/14 foundation model, with all training data generated exclusively via the IDPERTURB synthetic face generation framework and without any use of real face images. Two tracks were organized to assess adaptation under large-scale and limited-data conditions, attracting 28 registered teams and four valid submitting teams from academic and industrial institutions across Asia, Africa, Europe, North and South America.

The evaluation results demonstrate that adapting CLIP with synthetic face data substantially improves recognition performance over the off-the-shelf model across all benchmarks. In the Full Data Track, DMSTI-Neurotechnology's full fine-tuning strategy with Sub-Center ArcFace achieves the strongest results in both recognition accuracy and demographic fairness, while in the Limited Data Track, Idiap-BSP's rsLoRA approach proves the most robust and consistent method, outperforming other submitted solution and the baseline FRoundation across all reported benchmarks. Fairness evaluation further reveals that, while fine-tuning significantly reduces demographic disparity relative to the pre-trained CLIP model for most submitted solutions, systematic performance gaps across ethnic groups persist, pointing to an important open challenge for the community. The competition protocol, evaluation benchmarks, and ranking methodology introduced in this work provide a reproducible framework for future comparisons in this emerging research direction. The findings motivate further investigation into the role of synthetic data diversity in shaping the generalization of adapted foundation models.

\section*{Acknowledgment}
This research work has been funded by the German Federal Ministry of Education and Research and the Hessen State Ministry for Higher Education, Research and the Arts within their joint support of the National Research Center for Applied Cybersecurity ATHENE.

{\small
\bibliographystyle{ieee}
\bibliography{egbib}

@String(CVPR= {IEEE Conf. Comput. Vis. Pattern Recog.})

@String(ICCV= {Int. Conf. Comput. Vis.})

@String(ECCV= {Eur. Conf. Comput. Vis.})

@String(ACCV  = {ACCV})

@String(ICLR = {Int. Conf. Learn. Represent.})

@String(CVPR  = {CVPR})

@String(ICCV  = {ICCV})

@String(ECCV  = {ECCV})

@String(ICLR  = {ICLR})

@inproceedings{DBLP:conf/iccv/DanLXD0XS23,
  author       = {Jun Dan and
                  Yang Liu and
                  Haoyu Xie and
                  Jiankang Deng and
                  Haoran Xie and
                  Xuansong Xie and
                  Baigui Sun},
  title        = {TransFace: Calibrating Transformer Training for Face Recognition from
                  a Data-Centric Perspective},
  booktitle    = {{ICCV}},
  pages        = {20585--20596},
  publisher    = {{IEEE}},
  year         = {2023}
}

@inproceedings{DBLP:conf/cvpr/DengGXZ19,
  author       = {Jiankang Deng and
                  Jia Guo and
                  Niannan Xue and
                  Stefanos Zafeiriou},
  title        = {ArcFace: Additive Angular Margin Loss for Deep Face Recognition},
  booktitle    = {{CVPR}},
  pages        = {4690--4699},
  publisher    = {Computer Vision Foundation / {IEEE}},
  year         = {2019}
}

@inproceedings{wang2018cosfacelargemargincosine,
  author       = {Hao Wang and
                  Yitong Wang and
                  Zheng Zhou and
                  Xing Ji and
                  Dihong Gong and
                  Jingchao Zhou and
                  Zhifeng Li and
                  Wei Liu},
  title        = {CosFace: Large Margin Cosine Loss for Deep Face Recognition},
  booktitle    = {{CVPR}},
  pages        = {5265--5274},
  publisher    = {Computer Vision Foundation / {IEEE} Computer Society},
  year         = {2018}
}

@article{Deng_2022,
   title={ArcFace: Additive Angular Margin Loss for Deep Face Recognition},
   volume={44},
   ISSN={1939-3539},
   url={http://dx.doi.org/10.1109/TPAMI.2021.3087709},
   DOI={10.1109/tpami.2021.3087709},
   number={10},
   journal={IEEE Transactions on Pattern Analysis and Machine Intelligence},
   publisher={Institute of Electrical and Electronics Engineers (IEEE)},
   author={Deng, Jiankang and Guo, Jia and Yang, Jing and Xue, Niannan and Kotsia, Irene and Zafeiriou, Stefanos},
   year={2022},
   month=oct, pages={5962–5979} }

@inproceedings{huang:inria-00321923,
  TITLE = {{Labeled Faces in the Wild: A Database for Studying Face Recognition in Unconstrained Environments}},
  AUTHOR = {Huang, Gary B. and Mattar, Marwan and Berg, Tamara and Learned-Miller, Eric},
  URL = {https://inria.hal.science/inria-00321923},
  BOOKTITLE = {{Workshop on Faces in 'Real-Life' Images: Detection, Alignment, and Recognition}},
  ADDRESS = {Marseille, France},
  ORGANIZATION = {{Erik Learned-Miller and Andras Ferencz and Fr{\'e}d{\'e}ric Jurie}},
  YEAR = {2008},
  MONTH = Oct,
  HAL_ID = {inria-00321923},
  HAL_VERSION = {v1},
}

@article{DBLP:journals/corr/abs-1708-08197,
  author       = {Tianyue Zheng and
                  Weihong Deng and
                  Jiani Hu},
  title        = {Cross-Age {LFW:} {A} Database for Studying Cross-Age Face Recognition
                  in Unconstrained Environments},
  journal      = {CoRR},
  volume       = {abs/1708.08197},
  year         = {2017}
}

@inproceedings{c3517bca662f4193a58fd8f9e3145c8f,
title = "Frontal to profile face verification in the wild",
author = "Soumyadip Sengupta and Chen, {Jun Cheng} and Carlos Castillo and Patel, {Vishal M.} and Rama Chellappa and Jacobs, {David W.}",
note = "Publisher Copyright: {\textcopyright} 2016 IEEE.; IEEE Winter Conference on Applications of Computer Vision, WACV 2016 ; Conference date: 07-03-2016 Through 10-03-2016",
year = "2016",
month = may,
day = "23",
doi = "10.1109/WACV.2016.7477558",
language = "English (US)",
series = "2016 IEEE Winter Conference on Applications of Computer Vision, WACV 2016",
publisher = "Institute of Electrical and Electronics Engineers Inc.",
booktitle = "2016 IEEE Winter Conference on Applications of Computer Vision, WACV 2016",
}

@inproceedings{DBLP:conf/icb/MazeADKMO0NACG18,
  author       = {Brianna Maze and
                  Jocelyn C. Adams and
                  James A. Duncan and
                  Nathan D. Kalka and
                  Tim Miller and
                  Charles Otto and
                  Anil K. Jain and
                  W. Tyler Niggel and
                  Janet Anderson and
                  Jordan Cheney and
                  Patrick Grother},
  title        = {{IARPA} Janus Benchmark - {C:} Face Dataset and Protocol},
  booktitle    = {{ICB}},
  pages        = {158--165},
  publisher    = {{IEEE}},
  year         = {2018}
}

@inproceedings{inproceedingsijbb,
  author       = {Cameron Whitelam and
                  Emma Taborsky and
                  Austin Blanton and
                  Brianna Maze and
                  Jocelyn C. Adams and
                  Tim Miller and
                  Nathan D. Kalka and
                  Anil K. Jain and
                  James A. Duncan and
                  Kristen Allen and
                  Jordan Cheney and
                  Patrick Grother},
  title        = {{IARPA} Janus Benchmark-B Face Dataset},
  booktitle    = {{CVPR} Workshops},
  pages        = {592--600},
  publisher    = {{IEEE} Computer Society},
  year         = {2017}
}

@TechReport{CPLFWTech,
  author =       {T. Zheng and W. Deng},
  title =        {Cross-pose LFW: A database for studying cross-pose face recognition in unconstrained environments},
  institution =  {Beijing University of Posts and Telecommunications},
  year =         {2018},
  number =       {18-01},
  month =        {February}}

@inproceedings{moschoglou2017agedb, title={Agedb: the first manually collected, in-the-wild age database},
  author={Moschoglou, Stylianos and Papaioannou, Athanasios and Sagonas, Christos and Deng, Jiankang and Kotsia, Irene and Zafeiriou, Stefanos},
  booktitle={Proceedings of the IEEE Conference on Computer Vision and Pattern Recognition Workshop},
  volume={2},
  number={3},
  pages={5},
  year={2017}
}

@inproceedings{ElasticFace,
  author       = {Fadi Boutros and
                  Naser Damer and
                  Florian Kirchbuchner and
                  Arjan Kuijper},
  title        = {ElasticFace: Elastic Margin Loss for Deep Face Recognition},
  booktitle    = {{IEEE/CVF} Conference on Computer Vision and Pattern Recognition Workshops,
                  {CVPR} Workshops 2022, New Orleans, LA, USA, June 19-20, 2022},
  pages        = {1577--1586},
  publisher    = {{IEEE}},
  year         = {2022},
  url          = {https://doi.org/10.1109/CVPRW56347.2022.00164},
  doi          = {10.1109/CVPRW56347.2022.00164},
  bibsource    = {dblp computer science bibliography, https://dblp.org}
}

@inproceedings{guo2016ms,
  author    = {Yandong Guo and
               Lei Zhang and
               Yuxiao Hu and
               Xiaodong He and
               Jianfeng Gao},
  editor    = {Bastian Leibe and
               Jiri Matas and
               Nicu Sebe and
               Max Welling},
  title     = {MS-Celeb-1M: {A} Dataset and Benchmark for Large-Scale Face Recognition},
  booktitle = {Computer Vision - {ECCV} 2016 - 14th European Conference, Amsterdam,
               The Netherlands, October 11-14, 2016, Proceedings, Part {III}},
  series    = {Lecture Notes in Computer Science},
  volume    = {9907},
  pages     = {87--102},
  publisher = {Springer},
  year      = {2016},
  url       = {https://doi.org/10.1007/978-3-319-46487-9\_6},
  doi       = {10.1007/978-3-319-46487-9\_6},
  bibsource = {dblp computer science bibliography, https://dblp.org}
}

@inproceedings{DBLP:conf/accv/ChengZG18,
  author       = {Zhiyi Cheng and
                  Xiatian Zhu and
                  Shaogang Gong},
  title        = {Low-Resolution Face Recognition},
  booktitle    = {{ACCV} {(3)}},
  series       = {Lecture Notes in Computer Science},
  volume       = {11363},
  pages        = {605--621},
  publisher    = {Springer},
  year         = {2018}
}

@inproceedings{Randaugment_CVPR,
  author    = {Ekin D. Cubuk and
               Barret Zoph and
               Jonathon Shlens and
               Quoc V. Le},
  title     = {Randaugment: Practical automated data augmentation with a reduced
               search space},
  booktitle = {2020 {IEEE/CVF} Conference on Computer Vision and Pattern Recognition,
               {CVPR} Workshops 2020, Seattle, WA, USA, June 14-19, 2020},
  pages     = {3008--3017},
  publisher = {Computer Vision Foundation / {IEEE}},
  year      = {2020},
  url       = {https://openaccess.thecvf.com/content\_CVPRW\_2020/html/w40/Cubuk\_Randaugment\_Practical\_Automated\_Data\_Augmentation\_With\_a\_Reduced\_Search\_Space\_CVPRW\_2020\_paper.html},
  doi       = {10.1109/CVPRW50498.2020.00359},
  bibsource = {dblp computer science bibliography, https://dblp.org}
}

@article{DBLP:journals/ivc/ChettaouiDB25,
  author       = {Tahar Chettaoui and
                  Naser Damer and
                  Fadi Boutros},
  title        = {FRoundation: Are foundation models ready for face recognition?},
  journal      = {Image Vis. Comput.},
  volume       = {156},
  pages        = {105453},
  year         = {2025}
}

@inproceedings{DBLP:conf/icml/RadfordKHRGASAM21,
  author       = {Alec Radford and
                  Jong Wook Kim and
                  Chris Hallacy and
                  Aditya Ramesh and
                  Gabriel Goh and
                  Sandhini Agarwal and
                  Girish Sastry and
                  Amanda Askell and
                  Pamela Mishkin and
                  Jack Clark and
                  Gretchen Krueger and
                  Ilya Sutskever},
  title        = {Learning Transferable Visual Models From Natural Language Supervision},
  booktitle    = {{ICML}},
  series       = {Proceedings of Machine Learning Research},
  volume       = {139},
  pages        = {8748--8763},
  publisher    = {{PMLR}},
  year         = {2021}
}

@inproceedings{DBLP:conf/cvpr/Kim0L22,
  author       = {Minchul Kim and
                  Anil K. Jain and
                  Xiaoming Liu},
  title        = {AdaFace: Quality Adaptive Margin for Face Recognition},
  booktitle    = {{CVPR}},
  pages        = {18729--18738},
  publisher    = {{IEEE}},
  year         = {2022}
}

@inproceedings{DBLP:conf/fgr/CaoSXPZ18,
  author       = {Qiong Cao and
                  Li Shen and
                  Weidi Xie and
                  Omkar M. Parkhi and
                  Andrew Zisserman},
  title        = {VGGFace2: {A} Dataset for Recognising Faces across Pose and Age},
  booktitle    = {13th {IEEE} International Conference on Automatic Face {\&} Gesture
                  Recognition, {FG} 2018, Xi'an, China, May 15-19, 2018},
  pages        = {67--74},
  publisher    = {{IEEE} Computer Society},
  year         = {2018},
  url          = {https://doi.org/10.1109/FG.2018.00020},
  doi          = {10.1109/FG.2018.00020},
  bibsource    = {dblp computer science bibliography, https://dblp.org}
}

@inproceedings{zhu2021webface260mbenchmarkunveilingpower,
  author       = {Zheng Zhu and
                  Guan Huang and
                  Jiankang Deng and
                  Yun Ye and
                  Junjie Huang and
                  Xinze Chen and
                  Jiagang Zhu and
                  Tian Yang and
                  Jiwen Lu and
                  Dalong Du and
                  Jie Zhou},
  title        = {WebFace260M: {A} Benchmark Unveiling the Power of Million-Scale Deep
                  Face Recognition},
  booktitle    = {{CVPR}},
  pages        = {10492--10502},
  publisher    = {Computer Vision Foundation / {IEEE}},
  year         = {2021}
}

@inproceedings{hu2021loralowrankadaptationlarge,
  author       = {Edward J. Hu and
                  Yelong Shen and
                  Phillip Wallis and
                  Zeyuan Allen{-}Zhu and
                  Yuanzhi Li and
                  Shean Wang and
                  Lu Wang and
                  Weizhu Chen},
  title        = {LoRA: Low-Rank Adaptation of Large Language Models},
  booktitle    = {{ICLR}},
  publisher    = {OpenReview.net},
  year         = {2022}
}

@ARTICLE{shahreza2025foundation,
  author={Otroshi Shahreza, Hatef and Marcel, Sébastien},
  journal={IEEE Transactions on Information Forensics and Security}, 
  title={Foundation Models and Biometrics: A Survey and Outlook}, 
  year={2025},
  volume={20},
  number={},
  pages={9113-9138},
  doi={10.1109/TIFS.2025.3602233}}

@inproceedings{DBLP:conf/eccv/PapantoniouLMDKZ24,
  author       = {Foivos Paraperas Papantoniou and
                  Alexandros Lattas and
                  Stylianos Moschoglou and
                  Jiankang Deng and
                  Bernhard Kainz and
                  Stefanos Zafeiriou},
  title        = {Arc2Face: {A} Foundation Model for ID-Consistent Human Faces},
  booktitle    = {{ECCV} {(37)}},
  series       = {Lecture Notes in Computer Science},
  volume       = {15095},
  pages        = {241--261},
  publisher    = {Springer},
  year         = {2024}
}

@misc{lirias3838501,
  title={On synthetic data: a brief introduction for data protection law dummies},
  author={Fontanillo L{\'o}pez, C{\'e}sar Augusto and Elbi, Abdullah},
  year={2022},
  publisher={EU Law Blog},
howpublished = {\url{https://europeanlawblog.eu/2022/09/22/on-synthetic-data-a-brief-introduction-for-data-protection-law-dummies/
}}
}

@article{oquab2024dinov2learningrobustvisual,
  author       = {Maxime Oquab and
                  Timoth{\'{e}}e Darcet and
                  Th{\'{e}}o Moutakanni and
                  Huy V. Vo and
                  Marc Szafraniec and
                  Vasil Khalidov and
                  Pierre Fernandez and
                  Daniel Haziza and
                  Francisco Massa and
                  Alaaeldin El{-}Nouby and
                  Mido Assran and
                  Nicolas Ballas and
                  Wojciech Galuba and
                  Russell Howes and
                  Po{-}Yao Huang and
                  Shang{-}Wen Li and
                  Ishan Misra and
                  Michael Rabbat and
                  Vasu Sharma and
                  Gabriel Synnaeve and
                  Hu Xu and
                  Herv{\'{e}} J{\'{e}}gou and
                  Julien Mairal and
                  Patrick Labatut and
                  Armand Joulin and
                  Piotr Bojanowski},
  title        = {DINOv2: Learning Robust Visual Features without Supervision},
  journal      = {Trans. Mach. Learn. Res.},
  volume       = {2024},
  year         = {2024}
}

@article{kalajdzievski2023rank,
  title={A rank stabilization scaling factor for fine-tuning with lora},
  author={Kalajdzievski, Damjan},
  year={2023}
}

@inproceedings{DBLP:conf/eccv/DengGLGZ20,
  author       = {Jiankang Deng and
                  Jia Guo and
                  Tongliang Liu and
                  Mingming Gong and
                  Stefanos Zafeiriou},
  title        = {Sub-center ArcFace: Boosting Face Recognition by Large-Scale Noisy
                  Web Faces},
  booktitle    = {{ECCV} {(11)}},
  series       = {Lecture Notes in Computer Science},
  pages        = {741--757},
  publisher    = {Springer},
  year         = {2020}
}

@inproceedings{boutros2026idperturb,
  author    = {Fadi Boutros and
               Eduarda Caldeira and
               Tahar Chettaoui and
               Naser Damer},
  title     = {IDperturb: Enhancing Variation in Synthetic Face Generation via Angular Perturbation},
  booktitle = {Proceedings of the IEEE/CVF Conference on Computer Vision and Pattern Recognition (CVPR)},
  year      = {2026}
}

@InProceedings{Magistri_2026_CVPR,
    author    = {Magistri, Simone and Goswami, Dipam and Mistretta, Marco and Twardowski, Bart{\l}omiej and van de Weijer, Joost and Bagdanov, Andrew D.},
    title     = {IsoCLIP: Decomposing CLIP Projectors for Efficient Intra-modal Alignment},
    booktitle = {Proceedings of the IEEE/CVF Conference on Computer Vision and Pattern Recognition (CVPR)},
    month     = {June},
    year      = {2026},
    pages     = {29315-29324}
}

@inproceedings{DBLP:conf/nips/TouvronVDJ19,
  author       = {Hugo Touvron and
                  Andrea Vedaldi and
                  Matthijs Douze and
                  Herv{\'{e}} J{\'{e}}gou},
  title        = {Fixing the train-test resolution discrepancy},
  booktitle    = {NeurIPS},
  pages        = {8250--8260},
  year         = {2019}
}

@inproceedings{DBLP:conf/iccv/WangDHTH19,
  author       = {Mei Wang and
                  Weihong Deng and
                  Jiani Hu and
                  Xunqiang Tao and
                  Yaohai Huang},
  title        = {Racial Faces in the Wild: Reducing Racial Bias by Information Maximization
                  Adaptation Network},
  booktitle    = {{ICCV}},
  pages        = {692--702},
  publisher    = {{IEEE}},
  year         = {2019}
}

@article{PAMI_BIAS,
  author       = {Mei Wang and
                  Yaobin Zhang and
                  Weihong Deng},
  title        = {Meta Balanced Network for Fair Face Recognition},
  journal      = {{IEEE} Trans. Pattern Anal. Mach. Intell.},
  volume       = {44},
  number       = {11},
  pages        = {8433--8448},
  year         = {2022},
  url          = {https://doi.org/10.1109/TPAMI.2021.3103191},
  doi          = {10.1109/TPAMI.2021.3103191},
  bibsource    = {dblp computer science bibliography, https://dblp.org}
}

@inproceedings{DBLP:journals/corr/abs-1911-10692,
  title={Mitigating bias in face recognition using skewness-aware reinforcement learning},
  author={Wang, Mei and Deng, Weihong},
  booktitle={Proceedings of the IEEE/CVF conference on computer vision and pattern recognition},
  pages={9322--9331},
  year={2020}
}

@article{DBLP:journals/inffus/MelziTVKRLDMFOZZYZWLTKZDBVGFFMUG24,
  author       = {Pietro Melzi and
                  Ruben Tolosana and
                  Rub{\'{e}}n Vera{-}Rodr{\'{\i}}guez and
                  Minchul Kim and
                  Christian Rathgeb and
                  Xiaoming Liu and
                  Ivan DeAndres{-}Tame and
                  Aythami Morales and
                  Julian Fi{\'{e}}rrez and
                  Javier Ortega{-}Garcia and
                  Weisong Zhao and
                  Xiangyu Zhu and
                  Zheyu Yan and
                  Xiaoyu Zhang and
                  Jinlin Wu and
                  Zhen Lei and
                  Suvidha Tripathi and
                  Mahak Kothari and
                  Md Haider Zama and
                  Debayan Deb and
                  Bernardo Biesseck and
                  Pedro Vidal and
                  Roger Granada and
                  Guilherme P. Fickel and
                  Gustavo F{\"{u}}hr and
                  David Menotti and
                  Alexander Unnervik and
                  Anjith George and
                  Christophe Ecabert and
                  Hatef Otroshi{-}Shahreza and
                  Parsa Rahimi and
                  S{\'{e}}bastien Marcel and
                  Ioannis Sarridis and
                  Christos Koutlis and
                  Georgia Baltsou and
                  Symeon Papadopoulos and
                  Christos Diou and
                  Nicol{\`{o}} Di Domenico and
                  Guido Borghi and
                  Lorenzo Pellegrini and
                  Enrique Mas{-}Candela and
                  {\'{A}}ngela S{\'{a}}nchez{-}P{\'{e}}rez and
                  Andrea Atzori and
                  Fadi Boutros and
                  Naser Damer and
                  Gianni Fenu and
                  Mirko Marras},
  title        = {FRCSyn-onGoing: Benchmarking and comprehensive evaluation of real
                  and synthetic data to improve face recognition systems},
  journal      = {Inf. Fusion},
  volume       = {107},
  pages        = {102322},
  year         = {2024}
}

@inproceedings{DBLP:conf/fgr/Otroshi-Shahreza24,
  author       = {Hatef Otroshi{-}Shahreza and
                  Christophe Ecabert and
                  Anjith George and
                  Alexander Unnervik and
                  S{\'{e}}bastien Marcel and
                  Nicol{\`{o}} Di Domenico and
                  Guido Borghi and
                  Davide Maltoni and
                  Fadi Boutros and
                  Julia Vogel and
                  Naser Damer and
                  {\'{A}}ngela S{\'{a}}nchez{-}P{\'{e}}rez and
                  Enrique Mas{-}Candela and
                  Jorge Calvo{-}Zaragoza and
                  Bernardo Biesseck and
                  Pedro Vidal and
                  Roger Granada and
                  David Menotti and
                  Ivan DeAndres{-}Tame and
                  Simone Maurizio La Cava and
                  Sara Concas and
                  Pietro Melzi and
                  Ruben Tolosana and
                  Rub{\'{e}}n Vera{-}Rodr{\'{\i}}guez and
                  Gianpaolo Perelli and
                  Giulia Orr{\`{u}} and
                  Gian Luca Marcialis and
                  Julian Fi{\'{e}}rrez},
  title        = {{SDFR:} Synthetic Data for Face Recognition Competition},
  booktitle    = {{FG}},
  pages        = {1--9},
  publisher    = {{IEEE}},
  year         = {2024}
}

@article{DEANDRESTAME2025103099,
title = {Second FRCSyn-onGoing: Winning solutions and post-challenge analysis to improve face recognition with synthetic data},
journal = {Information Fusion},
pages = {103099},
year = {2025},
issn = {1566-2535},
doi = {https://doi.org/10.1016/j.inffus.2025.103099},
url = {https://www.sciencedirect.com/science/article/pii/S1566253525001721},
author = {Ivan DeAndres-Tame and Ruben Tolosana and Pietro Melzi and Ruben Vera-Rodriguez and Minchul Kim and Christian Rathgeb and Xiaoming Liu and Luis F. Gomez and Aythami Morales and Julian Fierrez and Javier Ortega-Garcia and Zhizhou Zhong and Yuge Huang and Yuxi Mi and Shouhong Ding and Shuigeng Zhou and Shuai He and Lingzhi Fu and Heng Cong and Rongyu Zhang and Zhihong Xiao and Evgeny Smirnov and Anton Pimenov and Aleksei Grigorev and Denis Timoshenko and Kaleb Mesfin Asfaw and Cheng Yaw Low and Hao Liu and Chuyi Wang and Qing Zuo and Zhixiang He and Hatef Otroshi Shahreza and Anjith George and Alexander Unnervik and Parsa Rahimi and Sébastien Marcel and Pedro C. Neto and Marco Huber and Jan Niklas Kolf and Naser Damer and Fadi Boutros and Jaime S. Cardoso and Ana F. Sequeira and Andrea Atzori and Gianni Fenu and Mirko Marras and Vitomir Štruc and Jiang Yu and Zhangjie Li and Jichun Li and Weisong Zhao and Zhen Lei and Xiangyu Zhu and Xiao-Yu Zhang and Bernardo Biesseck and Pedro Vidal and Luiz Coelho and Roger Granada and David Menotti}
}

@misc{gdpr,
    author={{The European Parliament and the Council of the European Union}},
    title={General Data Protection Regulation},
    year={2016}
}

@article{SyntheticFRSurvay,
  author       = {Fadi Boutros and
                  Vitomir Struc and
                  Julian Fi{\'{e}}rrez and
                  Naser Damer},
  title        = {Synthetic data for face recognition: Current state and future prospects},
  journal      = {Image Vis. Comput.},
  volume       = {135},
  pages        = {104688},
  year         = {2023},
  url          = {https://doi.org/10.1016/j.imavis.2023.104688},
  doi          = {10.1016/j.imavis.2023.104688},
  bibsource    = {dblp computer science bibliography, https://dblp.org}
}

@inproceedings{DBLP:conf/cvpr/Kim00023,
  author       = {Minchul Kim and
                  Feng Liu and
                  Anil K. Jain and
                  Xiaoming Liu},
  title        = {DCFace: Synthetic Face Generation with Dual Condition Diffusion Model},
  booktitle    = {{CVPR}},
  pages        = {12715--12725},
  publisher    = {{IEEE}},
  year         = {2023}
}

@inproceedings{DBLP:conf/iccv/BoutrosGKD23,
  author       = {Fadi Boutros and
                  Jonas Henry Grebe and
                  Arjan Kuijper and
                  Naser Damer},
  title        = {IDiff-Face: Synthetic-based Face Recognition through Fizzy Identity-Conditioned
                  Diffusion Models},
  booktitle    = {{ICCV}},
  pages        = {19593--19604},
  publisher    = {{IEEE}},
  year         = {2023}
}

@article{DBLP:journals/tbbis/BoutrosHLSD24,
  author       = {Fadi Boutros and
                  Marco Huber and
                  Anh Thi Luu and
                  Patrick Siebke and
                  Naser Damer},
  title        = {SFace2: Synthetic-Based Face Recognition With w-Space Identity-Driven
                  Sampling},
  journal      = {{IEEE} Trans. Biom. Behav. Identity Sci.},
  volume       = {6},
  number       = {3},
  pages        = {290--303},
  year         = {2024}
}

@article{DBLP:journals/corr/YiLLL14a,
  author       = {Dong Yi and
                  Zhen Lei and
                  Shengcai Liao and
                  Stan Z. Li},
  title        = {Learning Face Representation from Scratch},
  journal      = {CoRR},
  volume       = {abs/1411.7923},
  year         = {2014}
}

@inproceedings{DBLP:conf/wacv/MelziTVKRLDMFOZZYZWLTKZDBVGFFMUGEORMSK24,
  author       = {Pietro Melzi and
                  Ruben Tolosana and
                  Rub{\'{e}}n Vera{-}Rodr{\'{\i}}guez and
                  Minchul Kim and
                  Christian Rathgeb and
                  Xiaoming Liu and
                  Ivan DeAndres{-}Tame and
                  Aythami Morales and
                  Julian Fi{\'{e}}rrez and
                  Javier Ortega{-}Garcia and
                  Weisong Zhao and
                  Xiangyu Zhu and
                  Zheyu Yan and
                  Xiaoyu Zhang and
                  Jinlin Wu and
                  Zhen Lei and
                  Suvidha Tripathi and
                  Mahak Kothari and
                  Md Haider Zama and
                  Debayan Deb and
                  Bernardo Biesseck and
                  Pedro Vidal and
                  Roger Granada and
                  Guilherme P. Fickel and
                  Gustavo F{\"{u}}hr and
                  David Menotti and
                  Alexander Unnervik and
                  Anjith George and
                  Christophe Ecabert and
                  Hatef Otroshi{-}Shahreza and
                  Parsa Rahimi and
                  S{\'{e}}bastien Marcel and
                  Ioannis Sarridis and
                  Christos Koutlis and
                  Georgia Baltsou and
                  Symeon Papadopoulos and
                  Christos Diou and
                  Nicol{\`{o}} Di Domenico and
                  Guido Borghi and
                  Lorenzo Pellegrini and
                  Enrique Mas{-}Candela and
                  {\'{A}}ngela S{\'{a}}nchez{-}P{\'{e}}rez and
                  Andrea Atzori and
                  Gianni Fenu and
                  Fadi Boutros and
                  Mirko Marras and
                  Naser Damer},
  title        = {FRCSyn Challenge at {WACV} 2024: Face Recognition Challenge in the
                  Era of Synthetic Data},
  booktitle    = {{WACV} (Workshops)},
  pages        = {892--901},
  publisher    = {{IEEE}},
  year         = {2024}
}

@article{DBLP:journals/tmlr/BidermanPOPGJKH24,
  author       = {Dan Biderman and
                  Jacob P. Portes and
                  Jose Javier Gonzalez Ortiz and
                  Mansheej Paul and
                  Philip Greengard and
                  Connor Jennings and
                  Daniel King and
                  Sam Havens and
                  Vitaliy Chiley and
                  Jonathan Frankle and
                  Cody Blakeney and
                  John Patrick Cunningham},
  title        = {LoRA Learns Less and Forgets Less},
  journal      = {Trans. Mach. Learn. Res.},
  volume       = {2024},
  year         = {2024}
}

@article{borsukiewicz2026beyond,
  title={Beyond real faces: synthetic datasets can achieve reliable recognition performance without privacy compromise},
  author={Borsukiewicz, Pawe{\l} and Boutros, Fadi and Olatunji, Iyiola E and Beumier, Charles and Ouedraogo, Wendk{\^u}uni C and Klein, Jacques and Bissyand{\'e}, Tegawend{\'e} F},
  journal={npj Artificial Intelligence},
  year={2026},
  publisher={Nature Publishing Group}
}
}

\end{document}